\documentclass[sigconf]{acmart}
\usepackage{float}  
\usepackage{graphicx}
\usepackage{subcaption}
\usepackage{cleveref}
\usepackage{booktabs}
\usepackage{multirow}
\usepackage{caption}

\AtBeginDocument{%
  \providecommand\BibTeX{{%
    \normalfont B\kern-0.5em{\scshape i\kern-0.25em b}\kern-0.8em\TeX}}}

\setcopyright{acmlicensed}
\copyrightyear{2018}
\acmYear{2018}
\acmDOI{XXXXXXX.XXXXXXX}

\acmConference[MM'24]{Make sure to enter the correct
  conference title from your rights confirmation email}{October 28 - November 1,
  2024}{Melbourne, Australia.}
\acmISBN{978-1-4503-XXXX-X/18/06}

\begin{document}

\title{Exploring Robust Face-Voice Matching in Multilingual Environments}


\author{Jiehui Tang}
\authornote{Equally contribution}

\email{hfut2021214635@gmail.com}

\affiliation{%
  \institution{Hefei University of Technology}
  \city{Hefei}
  \state{Anhui}
  \country{China}}

\author{Xiaofei Wang}
\authornotemark[1]

\email{xiaofei2027@gmail.com}

\affiliation{%
  \institution{Hefei University of Technology}
  \city{Hefei}
  \state{Anhui}
  \country{China}}

\author{Zhen Xiao}
\authornotemark[1]

\email{xiaozhen0305@hotmail.com}

\affiliation{%
  \institution{Hefei University of Technology}
  \city{Hefei}
  \state{Anhui}
  \country{China}}

\author{Jiayi Liu}

\email{liujiayi257124@stu.vma.edu.cn }

\affiliation{%
  \institution{Vanke Meisha Academy}
  \city{Shenzhen}
  \state{Guangdong}
  \country{China}}

\author{Xueliang Liu}
\email{liuxueliang1982@gmail.com}
\authornote{Corresponding authors}

\affiliation{%
  \institution{Hefei University of Technology \\ Hefei Comprehensive National Science Center, Institute of Artificial Intelligence}
  \city{Hefei}
  \state{Anhui}
  \country{China}}

\author{Richang Hong}
\email{hongrc.hfut@gmail.com}

\affiliation{%
  \institution{Hefei University of Technology}
  \city{Hefei}
  \state{Anhui}
  \country{China}}

\renewcommand{\shortauthors}{Jiehui Tang, Xiaofei Wang, Zhen Xiao, Jiayi Liu, Xueliang Liu, \& Richang Hong}

\begin{abstract}
  This paper presents Team Xaiofei's innovative approach to exploring Face-Voice Association in Multilingual Environments (FAME) at ACM Multimedia 2024. We focus on the impact of different languages in face-voice matching by building upon Fusion and Orthogonal Projection (FOP), introducing four key components: a dual-branch structure, dynamic sample pair weighting, robust data augmentation, and score polarization strategy. Our dual-branch structure serves as an auxiliary mechanism to better integrate and provide more comprehensive information. We also introduce a dynamic weighting mechanism for various sample pairs to optimize learning. Data augmentation techniques are employed to enhance the model’s generalization across diverse conditions. Additionally, score polarization strategy based on age and gender matching confidence clarifies and accentuates the final results. Our methods demonstrate significant effectiveness, achieving an equal error rate (EER) of 20.07 on the V2-EH dataset and 21.76 on the V1-EU dataset.

\end{abstract}





\maketitle
\begin{figure}[!t]
    \centering
    \includegraphics[width=\linewidth]{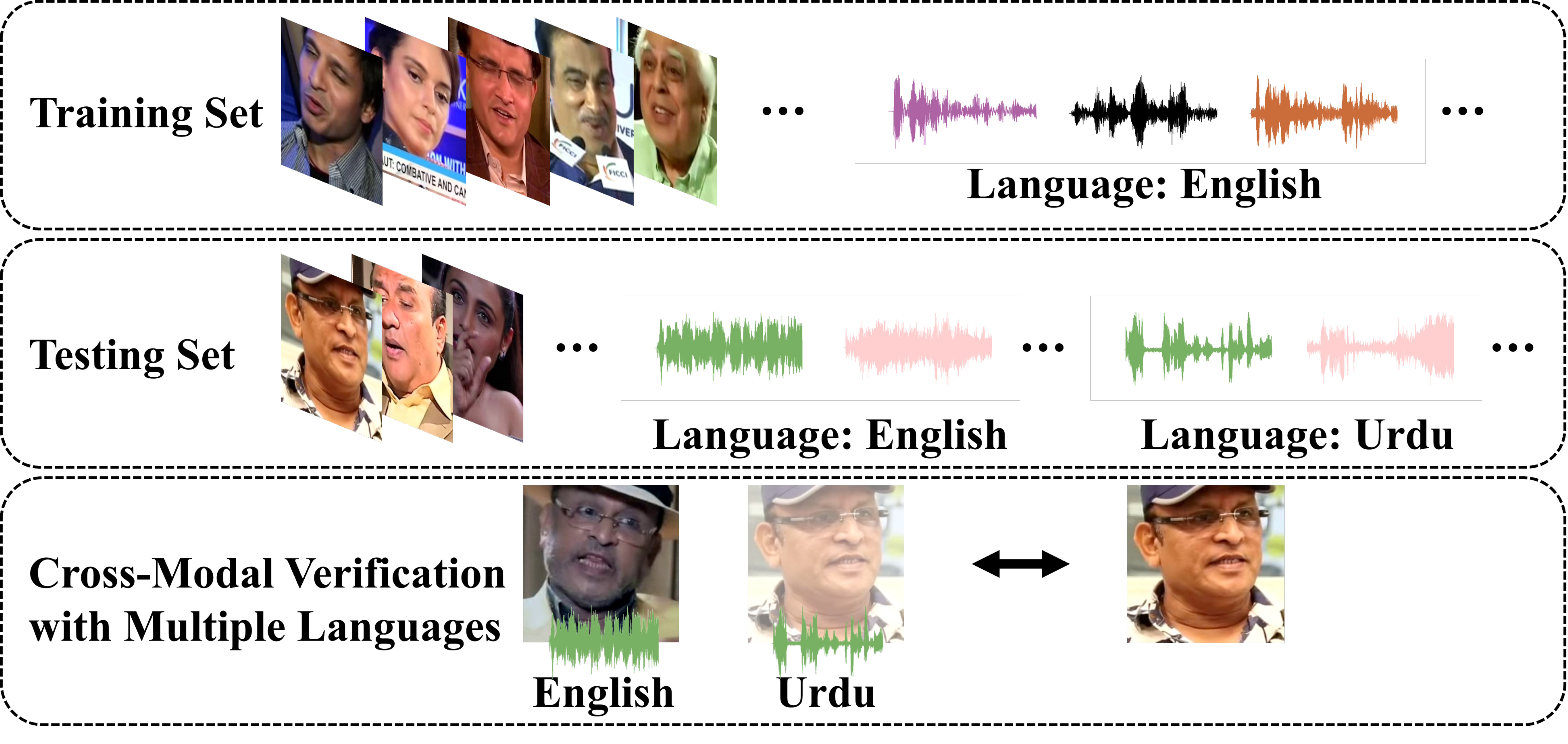}
    \caption{Illustration of the dataset and cross-modal verification process. The training set contains audio and facial features in English (Urdu). The testing set includes samples in both English and Urdu. Cross-modal verification is performed using multiple languages, comparing facial features with audio features from different languages.}
    \Description{Illustration showing the dataset and cross-modal verification process with training and testing sets in English and Urdu, and verification using multiple languages.}
    \label{fig:1main}
    \vspace{-15pt} 
\end{figure}
\vspace{-5pt}  

\section{Introduction}
The association between faces and voices is a significant topic within the field of pattern recognition research, having garnered widespread attention in recent years \cite{10.1007/978-3-030-01261-8_5,wen2021seeking,8794873}. Faces and voices, as crucial biometric features, carry abundant identity information and complement each other. Therefore, the study of face-voice association holds significant importance.

The FAME 2024 challenge, hosted at ACM Multimedia 2024, builds upon the MAV-Celeb dataset \cite{Nawaz2021CrossModal}. It requires researchers to assess how multiple languages affect cross-modal verification. The objective of this verification task is to determine whether the face and voice in a given sample belong to the same individual. For this challenge, we use Fusion and Orthogonal Projection \cite{saeed2022fusion} as our baseline.

To achieve this, we incorporate targeted data augmentation techniques based on the training dataset and employ a dual-branch structure, enabling the model to learn richer information from diverse features. We believe that by increasing training sample pairs in different environmental scenarios, more effective differentiation can be achieved. Moreover, we use dynamic sample pair weighting to direct the model's attention to challenging sample pairs. For the final L2 results, we adjust the scores based on age and gender matching confidence, using a threshold and polarization factor to achieve more accurate outcomes.

In summary, our main contributions are as follows:
\begin{itemize}
\item We effectively leverage the dual-branch structure to learn from diverse features, thereby reducing the equal error rate (EER) of the model.
\item We implement a data augmentation strategy during model training, significantly enhancing the dataset's potential by randomly generating additional training samples.
\item We refine the loss function by dynamically assigning weights to pairs of training samples, thereby improving our approach to handling complex classification scenarios.
\item For the cross-modal verification task, our solution achieves top 3 results on the leaderboard with equal error rates (EER) of 20.07 and 21.76, respectively.
\end{itemize}
The remainder of this paper is structured as follows. Section 2 reviews related work, Section 3 details our methods and optimization strategies, Section 4 describes the implementation and experimental evaluation, and Section 5 offers conclusions from our research.
\section{RELATED WORK}
In recent years, interest in human voice-face cross-modal learning tasks surges, and studies demonstrate the feasibility of recognizing a person’s face from their voice or identifying a speaker solely from their face image. SVHF \cite{8578977} is the first to propose a machine learning algorithm for voice-face association learning, focusing on matching voices and faces. It uses a two-stream architecture to independently learn audio and visual features, which are then combined for cross-modal matching. Shah et al. \cite{10136626} propose a two-branch network that utilizes pre-trained models to extract joint representations from faces and voices, optimizing the network through categorical cross-entropy loss and validating that facial information enhances speaker recognition. Horiguchi et al. \cite{10.1145/3240508.3240601} propose a face-voice matching model with N-pair loss, significantly improving verification accuracy. Nagrani et al. \cite{10.1007/978-3-030-01261-8_5} enhance cross-modal embeddings with a curriculum mining strategy and contrastive loss.

FOP~\cite{saeed2022fusion} leverages complementary cues from both modalities to form enriched fused embeddings and uses orthogonality constraints to cluster them based on identity labels, enhancing cross-modal face-voice association tasks. DIMNet \cite{wen2018disjoint} uses attribute covariates as supervisory signals, improving cross-modal matching with multi-task learning. SSNet \cite{8945863} uses a single-stream network and novel loss function to map audio and visual signals into a shared space, eliminating the need for pairwise or triplet supervision.

SBNet \cite{10097207} introduces a novel single-branch network capable of learning discriminative representations for both unimodal and multimodal tasks, eliminating the need for separate networks and achieving superior performance in cross-modal verification and matching tasks. EmNet \cite{8794873} employs a distance learning framework with an elastic matching network and distance-based loss, handling binary and multi-way matching tasks. AML \cite{9320535} combines adversarial learning and metric learning to create modality-independent feature representations, reducing cross-modal discrepancies.

To achieve modality alignment and minimize semantic gaps between voices and faces, Wen et al. \cite{wen2021seeking} propose an adaptive learning framework with dynamic identity weights. Ning et al. \cite{9400909} use disentanglement techniques to separate modality-specific and invariant features, enhancing matching accuracy and robustness. CMPC \cite{ijcai2022p526} uses contrastive methods and unsupervised clustering to create semantic-positive pairs, improving robustness and accuracy by recalibrating contributions of unlearnable instances. Kim et al. \cite{kim2019learning} study the associations between human faces and voices by modeling the overlapping information between the two modalities, achieving performance similar to humans in matching tasks. Finally, Cheng et al. \cite{10.1145/3394171.3413710} employ an adversarial deep semantic matching network with generator and discriminator subnetworks, using triplet loss and multimodal center loss to enhance voice-face interactions and associations.
\section{METHOD}
Our method consists of four main components: a dual-branch structure, sample pair weight configuration, data augmentation, and score polarization strategy.
\subsection{Dual-branch Structure}
\begin{figure*}[t!]
\flushleft
\begin{subfigure}[b]{0.55\linewidth}
    \flushleft
    \includegraphics[height=5.3cm]{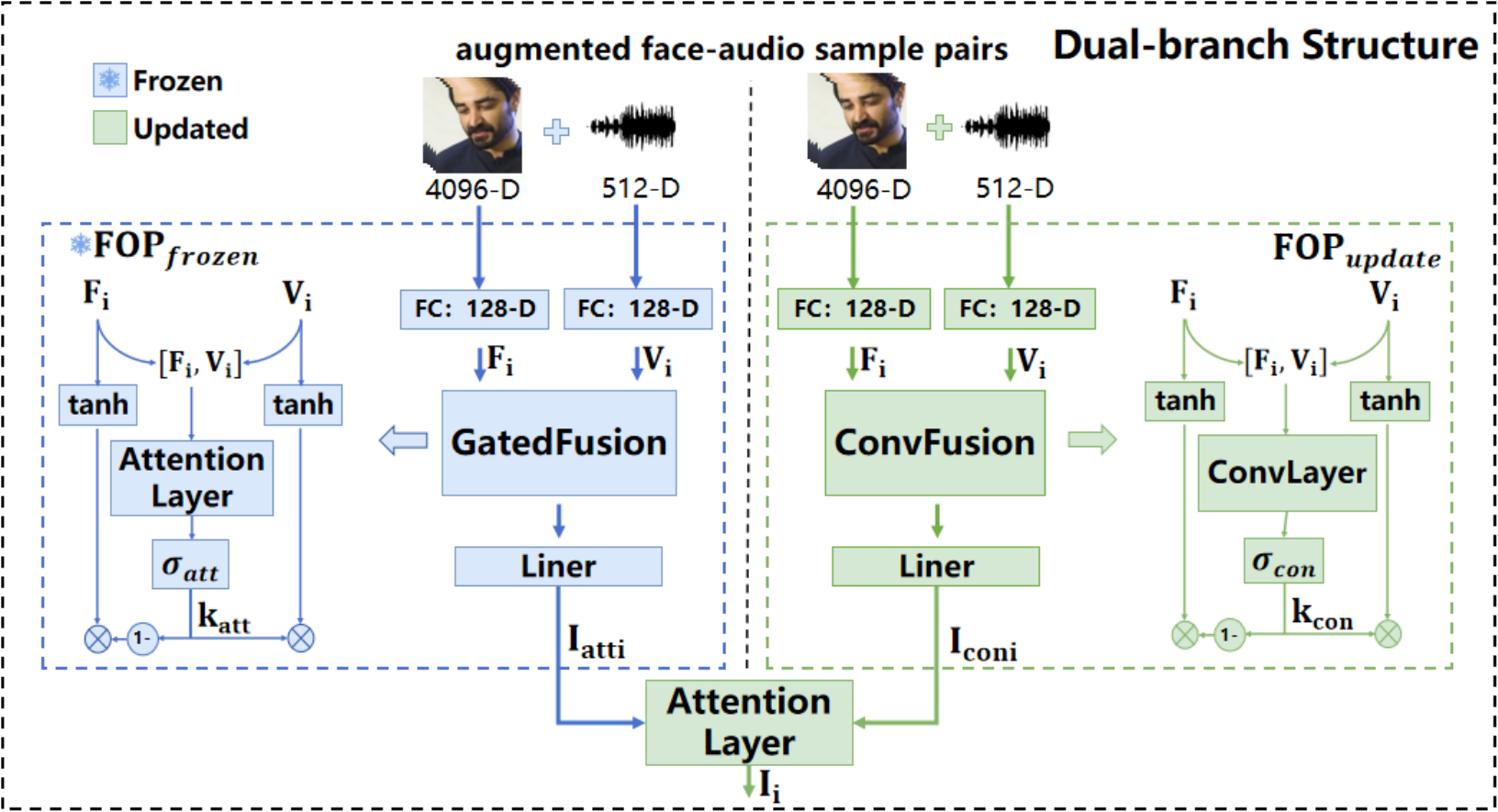}
    \caption{Dual-branch Structure}
    \label{fig:2a}
\end{subfigure}\hspace{-1em}
\begin{subfigure}[b]{0.23\linewidth}
    \centering
    \includegraphics[height=5.3cm]{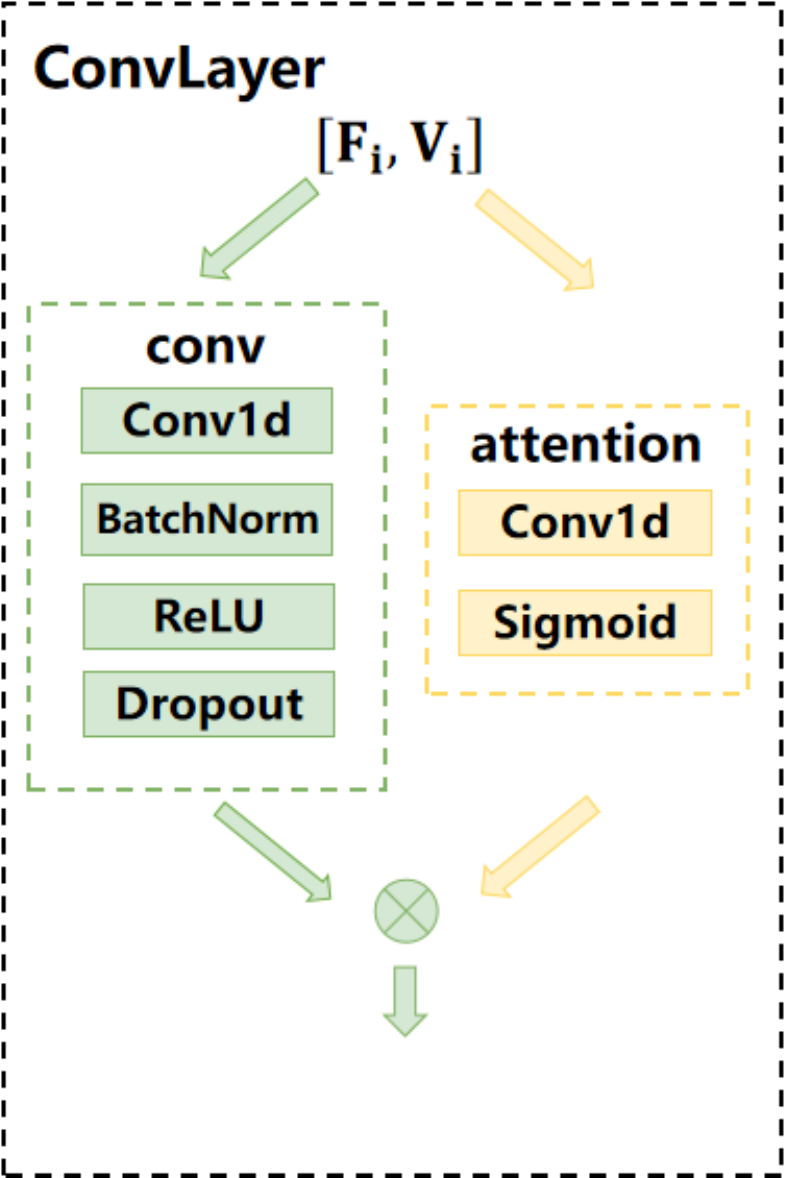}
    \caption{ConvLayer}
    \label{fig:2b}
\end{subfigure}\hspace{-1em}
\begin{subfigure}[b]{0.235\linewidth}
    \flushright
    \includegraphics[height=5.3cm]{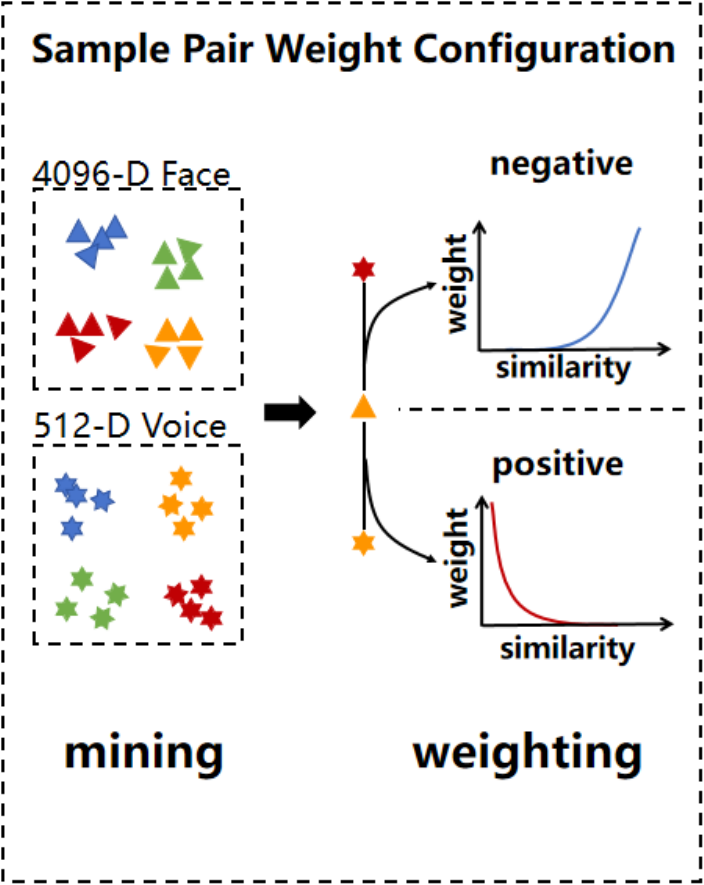}
    \caption{Weight Configuration}
    \label{fig:2c}
\end{subfigure}
\caption{The primary methodology employed by our team in the challenge: (a) the overarching architecture of the dual-branch model, (b) our convLayer for fusion (c) the dynamic weight configuration.}
\label{fig:2main}
\end{figure*}

In this section, we present the proposed model architecture as shown in Figure \ref{fig:2main} and provide a detailed description below.

To enable the model to learn more comprehensive information, we adopt a dual-branch structure to better integrate the data. This structure consists of the pre-trained $FOP_{\text{frozen}}$ and the $FOP_{\text{update}}$, which requires updates. The output results of the two feature output layers are fused through a learnable attention layer. Specifically, the $FOP_{\text{frozen}}$ is based on the previous work~\cite{saeed2022fusion}. We keep the parameters of the $FOP_{\text{frozen}}$ unchanged during training and train the $FOP_{\text{update}}$ from scratch. This design enables the $FOP_{\text{frozen}}$ model to act as a fixed feature extractor, while the $FOP_{\text{update}}$ actively complements the outputs of the $FOP_{\text{frozen}}$.

In our $FOP_{\text{update}}$, we adopt a novel fusion strategy to enhance the efficacy of multimodal feature fusion. Compared to the baseline model, we introduced a $ConvLayer$ for fusion in the process of multimodal fusion. Our designed $ConvLayer$ processes the input embeddings through two sequential operations: first, a convolution module for the initial transformation and extraction of features, followed by an attention module for generating the importance weights of the features.
Specifically, as shown in \Cref{fig:2b}, the input to the $ConvLayer$ consists of fused embeddings from audio and facial features. Initially, the $ConvLayer$ processes the input embeddings through a 1-D convolution layer, followed by batch normalization and ReLU activation to ensure nonlinearity and standardization of the features. Subsequently, these processed features pass through another 1-D convolution layer and a Sigmoid activation function to calculate the importance weights for each feature, corresponding to the convolution scores ($k_{\text{con}}$). Upon obtaining these importance weights, the module utilizes them to weight the feature maps, enhancing the focus on useful information and suppressing less important features, thereby facilitating more effective information fusion. Finally, this weighted feature map is compressed and output, completing a forward pass. This enhanced design allows the model to handle multimodal data with greater precision, improving its ability to capture key information. The specific operation is as follows:
\begin{equation}
k_{\text{con}} = \sigma_{\text{con}}(\text{conv}([F_{i}, V_{i}]) \odot \text{attention}([F_{i}, V_{i}]))
\end{equation}
where $\sigma_{con}$ denotes a Sigmoid operator, and the term $k_{\text{con}}$ represents the convolution scores obtained through the $ConvLayer$, which are processed via a Sigmoid activation function. In this context, $\text{conv}(\cdot)$ and $\text{attention}(\cdot)$ refer to the convolution module and the attention module, respectively. The symbol $\odot$ signifies element-wise multiplication. The variables $F_{i}$ and $V_{i}$ refer to the 128-dimensional face features and audio features, respectively, derived from the input augmented face-audio sample pairs through a linear transformation in a fully connected layer.

We improve the generalization and robustness by training the conventional $FOP_{\text{update}}$ using enhanced facial-audio samples. Additionally, we incorporate an attention layer to effectively modulate supplementary information. During the training phase, we compute attention scores to assess the interactions between the embeddings derived from the modal outputs of the dual branches. These scores are subsequently utilized to recalibrate the outputs of both branches, thereby optimizing the integration of their results. The precise method of fusion is detailed below:
\begin{equation}
I_i = W \odot I_{\text{coni}} + (1 - W) \odot I_{\text{atti}}
\end{equation}
where $I_{\text{coni}}$ denotes the fused embeddings from $FOP_{\text{update}}$, $I_{\text{atti}}$ denotes the fused embeddings from $FOP_{\text{frozen}}$, and $I_{\text{i}}$ represents the final fused embeddings of the dual-branch structure. Additionally, $W$ refers to the weights derived from an attention layer. This strategy not only enhances the model's ability to handle multimodal data but also improves performance stability in complex environments.
\subsection{Sample Pair Weight Configuration}
In our research, to effectively address the challenging scenarios of misclassifying positive sample pairs as different categories and negative sample pairs as the same category, we require the model to pay better attention to challenging samples. Therefore, we introduce a method \cite{Wang_2019_CVPR} for dynamically updating weights. Specifically, we dynamically adjust the weights of positive and negative sample pairs based on the similarity between the sample pairs (See \Cref{fig:2c}). Firstly, we calculate the similarity matrix $S$ between feature vectors:
\begin{equation}
S = \left( \frac{\sigma(WX + b)}{\|\sigma(WX + b)\|_2} \right) \left( \frac{\sigma(WX + b)}{\|\sigma(WX + b)\|_2} \right)^T
\end{equation}
here, $\frac{\sigma(WX + b)}{\|\sigma(WX + b)\|_2}$ represents the feature vector matrix normalized by the L2 norm.

For positive sample pairs, those with high similarity are assigned lower weights, whereas those with low similarity are assigned higher weights. For negative sample pairs, those with high similarity are assigned higher weights, and those with low similarity are assigned lower weights (or even zero). Subsequently, the dynamic weight loss for positive and negative sample pairs is calculated based on the similarity matrix:
\begin{equation}
L^+ = \sum e^{-\alpha \times (S_{ij} - \theta)}
\end{equation}
\begin{equation}
L^- = \sum e^{\beta \times (S_{ij} - \theta)}
\end{equation}
where $\alpha$ and $\beta$ represent the scaling factors for positive and negative sample pairs, respectively, and $\theta$ is the preset similarity threshold used to adjust the similarity in the loss function. $S_{ij}$ denotes the cosine similarity between the $i$-th and $j$-th samples. This weight adjustment aims to enhance the management of sample pairs that are difficult to distinguish. The overall loss $L$ is defined as:
\begin{equation}
L = \frac{1}{\alpha} \log(1 + L^+) + \frac{1}{\beta} \log(1 + L^-) + O
\end{equation}
\begin{equation}
O = (2.0 - \hat{S}^+) + 0.3 \times |\hat{S}^-|
\end{equation}
here, $\hat{S}^+$ and $\hat{S}^-$ represent the average similarities for positive and negative sample pairs, respectively, while $O$ represents the loss contribution from orthogonal projection, which is aimed at optimizing the spatial distribution of feature vectors to maximize the similarity of positive sample pairs and minimize that of negative pairs.

By constructing positive and negative sample pairs and weighting these pairs, our strategy aims to enhance the model's ability to correctly classify negative samples (i.e., those that should be dissimilar but are predicted to be similar) and to reduce misclassifications of positive samples (i.e., those that should be similar but are predicted to be dissimilar). Through this approach, we not only preserve the clustering advantages of the baseline model based on orthogonal projection but also significantly enhance the robustness and accuracy of the model in managing complex classification tasks. Particularly, the model demonstrates superior performance in identifying samples with complex features.
\subsection{Data Augmentation}
We observe that in the original dataset, data under the same identity label (i.e., the same speaker) usually includes multiple different scenes. However, in the original feature pairs, facial data from one scene corresponds exclusively to audio data from the same scene, forming one-to-one pairs. We believe that even in different scenes within the same identity label, audio and facial data can still form valid positive sample pairs. To enhance the robustness and diversity of our training set, we employ a data augmentation strategy aimed at disrupting the original pairing relationships within the same identity category.

Mathematically, let \( D = \{(f_i, a_i, l_i) \mid i = 1, 2, \ldots, N\} \) represent our original dataset, where \( f_i \) denotes the facial feature vector, \( a_i \) denotes the audio feature vector, and \( l_i \) denotes the identity label. In the original dataset, \( (f_i, a_i) \) pairs exist such that \( f_i \) and \( a_i \) are both from the same scene \( s \) under the same identity label \( l_i \).

Our augmentation strategy involves creating new pairs \( (f_i, a_j) \), where \( i \neq j \) and \( l_i = l_j \). This disruption introduces new pairings within the same identity label but from different scenes, simulating a broader array of real-world scenarios the model might encounter.

By randomizing these additional training sample pairs, we effectively simulate a broader range of scenarios, thereby improving the model's generalization ability. This approach can be formalized as follows:
\begin{align}
   \quad & P_{\text{original}} = \{(f_i, a_i) \mid i = 1, 2, \ldots, N\} \\
   \quad & P_{\text{augmented}} = \{(f_i, a_j) \mid i \neq j \text{ and } l_i = l_j\}
\end{align}
By augmenting the dataset in this manner, we enable the model to learn from diverse scenarios without the need for new data collection. This method maximizes the potential of the existing dataset.

\subsection{Score Polarization Strategy}
We utilize two pre-trained models~\cite{serengil2021lightface, burkhardt2023speech} trained on large datasets to predict the speaker's age and gender based on images and audio, respectively. These pre-trained models provide predicted age values ranging from 1 to 100 and gender prediction probabilities. Specifically, the predicted age values for audio and facial features are \(\hat{y}_a^A\) and \(\hat{y}_a^F\), respectively. The gender prediction probabilities (same gender) for audio and facial features are \(\hat{p}_g^A\) and \(\hat{p}_g^F\), respectively.

Matching confidence measures the similarity between audio and facial images in terms of age and gender. It plays a crucial role in adjusting the L2 score: the higher the matching confidence, the smaller the L2 score; conversely, the lower the matching confidence, the larger the L2 score.

Age matching confidence is calculated as the inverse of the absolute error between the predicted ages:

\begin{equation}
    C_a = \frac{1}{1 + |\hat{y}_a^A - \hat{y}_a^F|}
\end{equation}

Gender matching confidence is calculated using the gender prediction probabilities:

\begin{equation}
    C_g = \hat{p}_g^A \cdot \hat{p}_g^F + (1 - \hat{p}_g^A) \cdot (1 - \hat{p}_g^F)
\end{equation}

The overall matching confidence is a combination of age and gender matching confidence:

\begin{equation}
    C = w_a \cdot C_a + w_g \cdot C_g
\end{equation}

where \(w_a\) and \(w_g\) are the weights for age and gender matching confidence, respectively.

We adjust the L2 scores based on the matching confidence. A confidence threshold \(T\) and a polarization factor \(\alpha\) are set. If the confidence is higher than the threshold, the L2 score is reduced; otherwise, it is increased:

\begin{equation}
    \text{adjusted\_score} = 
    \begin{cases} 
        \frac{\text{initial\_score}}{\alpha} & \text{if } C > T \\
        \text{initial\_score} \cdot \alpha & \text{otherwise}
    \end{cases}
\end{equation}

Finally, we use evaluation methods to verify the adjustment effects and optimize the parameters based on the results. By adjusting the confidence threshold and polarization factor, the model performance can be further optimized.
\section{EXPERIMENTS}
\subsection{Dataset}
The FAME 2024 challenge aims to ascertain whether the voice and face data of participants in various scenarios correspond to the same individual under the influence of different languages. The FAME 2024 Challenge employs the MAV-Celeb dataset.
\begin{table}[!ht]
\centering
\caption{Data statistics of the MAV-Celeb dataset at the FAME 2024 Challenge} 
\label{tab:t1}
\begin{tabular}{ccccc} 
\hline
\toprule 
\multirow{2}{*}{Attributes} & \multicolumn{2}{c}{V1-EU} & \multicolumn{2}{c}{V2-EU} \\
\cmidrule(r){2-3} \cmidrule(r){4-5}
 & Train & Test & Train & Test \\
\midrule 
\#face-audio pairs & 18038 & 1513 & 19906 & 809 \\
\#identities & 64 & 6 & 78 & 6 \\
\bottomrule 
\hline
\end{tabular}
\vspace{-10pt}
\end{table}

As illustrated in \Cref{fig:1main}, the MAV-Celeb dataset is partitioned into training and testing divisions, featuring unique identities that communicate in the same language, in what is typically described as the unseen-unheard configuration. During the testing phase, the network is evaluated for its ability to process both familiar and entirely new languages as part of a cross-modal verification task involving multiple languages. Specifically, as shown in \Cref{tab:t1}, the dataset splits, V1-EU and V2-EH, include 64 and 78 identities for training, respectively, with each having 6 identities reserved for testing purposes. In the V1-EU split, there are a total of 18,038 face-audio pairs allocated for training and 1,513 for testing. In the V2-EU split, there are 19,906 face-audio pairs for training and 809 for testing.
\subsection{Experimental Setup}
\textbf{Implementation Details.} We follow the baseline \cite{saeed2022fusion} setup and conduct cross-modal verification and cross-modal matching tasks on V1-EU of the MAV-Celeb dataset. We utilize the embedding features of faces and voices extracted from VGGFace \cite{parkhi2015} and Utterance Level Aggregation \cite{xie2019} provided in the baseline. We exclude four identities from the training set to form a new test set for the experiments. The test files are generated through random cross-pairing, resulting in 1,432 test sets and 5,731 training sets in the English language environment, and 3,036 test sets and 12,145 training sets in the Urdu language environment. Subsequently, we conduct ablation experiments on various modules based on this dataset.

In the experiments, we employ a dual-branch model to evaluate the seen-heard and unseen-unheard identities. Our models are trained using the Adam optimizer with an initial learning rate of 1e-3 and a fixed learning rate strategy. The batch size is set at 64, and the embedding size at 128 dimensions. All experiments are conducted using the PyTorch framework on an NVIDIA RTX 3080Ti GPU.

\noindent \textbf{Evaluation Metric.} To assess the performance of the model, the Equal Error Rate (EER) is utilized as the evaluation metric, consistent with the evaluation plan of the FAME 2024 Challenge \cite{saeed2024fame}. The model outputs a score file, indicating the system's confidence in matching the face and voice, which determines whether they belong to the same person. For each test sample pair, consisting of a face feature $\mathbf{f}$ and a voice feature $\mathbf{v}$, the Euclidean distance $s$ between them is calculated as the score:
\begin{equation}
s = \sqrt{\sum_{i=1}^{n} (f_i - v_i)^2}
\end{equation}
where, $f_i$ and $v_i$ represent the $i$-th component of the face feature vector $\mathbf{f}$ and the voice feature vector $\mathbf{v}$, respectively, where $n$ denotes the dimension of the feature vectors. The higher the score, the more likely it is that the face and voice are from different individuals. We evaluate the performance through the scores to determine whether the sample pair is from the same person, and then calculate the EER.
\subsection{Ablation Study}
To evaluate the impact and performance of hyperparameter settings in each module, we conduct ablation studies on four different modules in this section. Notably, when examining an individual module, the other three modules are maintained at their default settings.
\begin{table}[!h] 
    \centering
    \caption{In the cross-modal validation task, ablation studies assess the dual-branch architecture and its fusion methods. ``Dual'' signifies the dual-branch setup. W, Att, and Conv each represent distinct fusion techniques. ``No Dual'' denotes the exclusion of the dual-branch structure, employing only the baseline and three additional modules.}
    \label{tab:t2}
    \begin{tabular}{@{}ccccc@{}} 
        \toprule
        \multirow{2}{*}{Method} & \multirow{2}{*}{Configuration} & \multicolumn{3}{c}{EER} \\ 
        \cmidrule{3-5}
        & & English & Urdu & \textbf{Avg} \\
        \midrule
        \multirow{2}{*}{Baseline \cite{saeed2022fusion}} & English train & 41.97 & 35.67 & \multirow{2}{*}{36.99} \\
                                  & Urdu train & 42.32 & 28.00 & \\
        \midrule
        \multirow{2}{*}{Dual+W}   & English train & 29.61 & 28.46 & \multirow{2}{*}{28.49} \\
                                  & Urdu train & 31.70 & 24.21 & \\
        \midrule
        \multirow{2}{*}{Dual+Att} & English train & 28.63 & 33.24 & \multirow{2}{*}{28.23} \\
                                  & Urdu train & 28.77 & 22.30 & \\
        \midrule
        \multirow{2}{*}{Dual+Conv} & English train & 33.10 & 32.87 & \multirow{2}{*}{30.76} \\
                                   & Urdu train & 33.94 & 23.12 & \\
        \midrule
        \multirow{2}{*}{No Dual}  & English train & 35.06 & 27.08 & \multirow{2}{*}{31.58} \\
                                   & Urdu train & 37.57 & 26.61 & \\
        \midrule
        \multirow{2}{*}{Our}  & English train & 28.56 & 33.23 & \multirow{2}{*}{\textbf{27.84}} \\
                                  & Urdu train & 28.63 & 20.95 & \\
        \bottomrule
    \end{tabular}
    \vspace{-10pt}
\end{table}
\begin{table}[!t]
    \centering
    \caption{In cross-modal validation tasks, ablation studies assess the impact of thresh $\theta$, the threshold parameter in the sample pair weight configuration. ``No Weighting'' indicates the absence of weight configuration, employing only the baseline and three additional modules.}
    \label{tab:t3}
    \begin{tabular}{@{}ccccc@{}}
        \toprule
        \multirow{2}{*}{Method} & \multirow{2}{*}{Configuration} & \multicolumn{3}{c}{EER} \\
        \cmidrule{3-5}
        & & English & Urdu & \textbf{Avg} \\
        \midrule
        \multirow{2}{*}{Baseline \cite{saeed2022fusion}} & English train & 41.97 & 35.67 & \multirow{2}{*}{36.99} \\
                                  & Urdu train & 42.32 & 28.00 & \\
        \midrule
        \multirow{2}{*}{thresh=0.4} & English train & 31.91 & 25.99 & \multirow{2}{*}{28.38} \\
                                  & Urdu train & 35.47 & 20.13 & \\
        \midrule
        \multirow{2}{*}{thresh=0.6} & English train & 28.63 & 33.24 & \multirow{2}{*}{28.23} \\
                                  & Urdu train & 28.77 & 22.30 & \\
        \midrule
        \multirow{2}{*}{thresh=0.8} & English train & 33.59 & 35.21 & \multirow{2}{*}{34.06} \\
                                   & Urdu train & 37.08 & 30.36 & \\
        \midrule
        \multirow{2}{*}{No Weighting} & English train & 33.94 & 29.05 & \multirow{2}{*}{31.22} \\
                                  & Urdu train & 34.57 & 27.31 & \\
        \midrule
        \multirow{2}{*}{Our}  & English train & 28.56 & 33.23 & \multirow{2}{*}{\textbf{27.84}} \\
                                  & Urdu train & 28.63 & 20.95 & \\
        \bottomrule
    \end{tabular}
    \vspace{-10pt}
\end{table}
\subsubsection{Dual-branch Structure} In this section, we conduct ablation experiments on the fusion methods of the dual-branch structure. We primarily design three types of fusion methods: the fusion of weights $W$, the fusion using attention structures, and the fusion through convolutional structures. We divide the configurations into three groups, each employing a different fusion method, while maintaining the default parameters for the other three modules, to assess the impact of various fusion methods on the dual-branch structure. Additionally, we use a baseline model and a comparison model that does not employ the dual-branch structure but includes the other three modules. The results, as shown in \Cref{tab:t2}, indicate that the inclusion of the dual-branch structure significantly enhances the performance. Furthermore, among the three fusion methods, the fusion method based on the attention structure achieves better scores compared to the others.
\subsubsection{Sample Pair Weight Configuration} In this section, we conduct ablation studies on thresh $\theta$, which is the most critical parameter within the sample pair weight configuration. This parameter is the preset similarity threshold used to adjust the similarity in the loss function, significantly influencing model performance. We set thresh at 0.4, 0.6, and 0.8, respectively, to assess the impact of each setting on the model. The experimental results, as shown in \Cref{tab:t3}, demonstrate that employing weight configuration effectively enhances model performance. Additionally, the model exhibits optimal performance when thresh is set at 0.6.
\subsubsection{Data Augmentation}
In this section, we conduct an ablation study on data augmentation. Specifically, we generate augmented data with different quantities, including 2x, 4x, and 6x the original data amount, and compare these with the unaugmented data. Through this data augmentation strategy, we aim to evaluate the impact of the additional data on model performance. The experimental results, as shown in Table \ref{tab:t4}, illustrate the model performance variations with different amounts of augmented data. The results indicate that the data augmentation strategy increases the diversity of the dataset by generating more sample pairs, which overall improves the model performance. Notably, when the data augmentation amount is 4x, the model performance is optimal.
\begin{table}[!t]
    \centering
    \caption{In the cross-modal validation task, ablation studies evaluate the effects of data augmentation on model performance, with datasets augmented to 2x, 4x, and 6x the original size.}
    \label{tab:t4}
    \begin{tabular}{@{}ccccc@{}}
        \toprule
        \multirow{2}{*}{Method} & \multirow{2}{*}{Configuration} & \multicolumn{3}{c}{EER} \\
        \cmidrule{3-5}
        & & English & Urdu & \textbf{Avg} \\
        \midrule
        \multirow{2}{*}{Baseline \cite{saeed2022fusion}} & English train & 41.97 & 35.67 & \multirow{2}{*}{36.99} \\
                                  & Urdu train & 42.32 & 28.00 & \\
        \midrule
        \multirow{2}{*}{Original Data} & English train & 33.59 & 30.80 & \multirow{2}{*}{31.47} \\
                                  & Urdu train & 30.80 & 34.19 & \\
        \midrule
        \multirow{2}{*}{2x Data} & English train & 32.33 & 28.69 & \multirow{2}{*}{29.03} \\
                                  & Urdu train & 30.45 & 24.67 & \\
        \midrule
        \multirow{2}{*}{4x Data} & English train & 32.19 & 30.73 & \multirow{2}{*}{28.23} \\
                                  & Urdu train & 34.01 & 21.97 & \\
        \midrule
        \multirow{2}{*}{6x Data} & English train & 32.12 & 28.95 & \multirow{2}{*}{29.20} \\
                                   & Urdu train & 32.40 & 23.32 & \\
        \midrule
        \multirow{2}{*}{Our}  & English train & 28.56 & 33.23 & \multirow{2}{*}{\textbf{27.84}} \\
                                  & Urdu train & 28.63 & 20.95 & \\
        \bottomrule
    \end{tabular}
    \vspace{-10pt}
\end{table}
\subsubsection{Score Polarization Strategy}
In this section, we conduct an ablation study on the score polarization strategy of the model. Specifically, we set different polarization factors $\alpha$, including 1, 1.1, 1.2, and 1.3. Through this adjustment strategy, we aim to assess the effect of different polarization factors on score polarization and the final model performance. The experimental results, as shown in the Table \ref{tab:t5}, illustrate the changes in model performance under different polarization factors. From the experimental results, it can be seen that adjusting the polarization factor improves model performance. Notably, the model performance is optimal when the polarization factor is set to 1.2.
\begin{table}[!t]
    \centering
    \caption{In the cross-modal validation task, ablation studies evaluate the effects of score polarization on model performance, with factors ranging from no polarization (1.0) to increasing levels at 1.1, 1.2, and 1.3.}
    \label{tab:t5}
    \begin{tabular}{@{}ccccc@{}}
       \toprule
        \multirow{2}{*}{Method} & \multirow{2}{*}{Configuration} & \multicolumn{3}{c}{EER} \\
        \cmidrule{3-5}
        & & English & Urdu & \textbf{Avg} \\
        \midrule
        \multirow{2}{*}{Baseline \cite{saeed2022fusion}} & English train & 41.97 & 35.67 & \multirow{2}{*}{36.99} \\
                                  & Urdu train & 42.32 & 28.00 & \\
        \midrule
        \multirow{2}{*}{factor=1.0} & English train & 35.68 & 32.44 & \multirow{2}{*}{34.32} \\
                                  & Urdu train & 42.60 & 26.55 & \\
        \midrule
        \multirow{2}{*}{factor=1.1} & English train & 31.91 & 30.63 & \multirow{2}{*}{30.00} \\
                                  & Urdu train & 34.71 & 22.73 & \\
        \midrule
        \multirow{2}{*}{factor=1.2} & English train & 32.19 & 30.73 & \multirow{2}{*}{28.23} \\
                                  & Urdu train & 34.01 & 21.97 & \\
        \midrule
        \multirow{2}{*}{factor=1.3} & English train & 31.77 & 30.63 & \multirow{2}{*}{29.52} \\
                                  & Urdu train & 33.66 & 22.00 & \\
        \midrule
        \multirow{2}{*}{Our}  & English train & 28.56 & 33.23 & \multirow{2}{*}{\textbf{27.84}} \\
                                  & Urdu train & 28.63 & 20.95 & \\
        \bottomrule
    \end{tabular}
    \vspace{-10pt}
\end{table}
\section{CONCLUSIONS}
In this paper, we introduce our solution developed for the FAME at ACM Multimedia 2024. Our approach utilizes a dual-branch structure and dynamic sample pair weight adjustment to enhance model performance. Additionally, we employ data augmentation and score polarization strategy based on the confidence of age and gender to improve prediction outcomes. The results indicate that our solution achieves a significantly better Equal Error Rate (EER) of \textbf{21.76} (rank \textbf{TOP 3}) on the V1-EU dataset of MAV-Celeb test set, substantially surpassing the baseline system (i.e., 33.4). In the future, we plan to address the Face-Voice Association from other perspectives. We will use DeepSpeech \cite{hannun2014deep} for audio feature extraction, employ a novel method \cite{ren2024monocular} for facial feature extraction, explore dynamic multimodal fusion techniques \cite{akbari2021vatt}, and integrate attention-based, correlation-aware multimodal fusion methods \cite{tsai2019multimodal}.
\begin{acks}
This work was supported by National Science and Technology Major Project (No. 2022ZD0118201),  National Natural Science Foundation of China (No. 62372151  and No.72188101). 

\end{acks}
\clearpage
\bibliographystyle{ACM-Reference-Format}
\bibliography{sample-base}


\begin{thebibliography}{26}


\ifx \showCODEN    \undefined \def \showCODEN     #1{\unskip}     \fi
\ifx \showDOI      \undefined \def \showDOI       #1{#1}\fi
\ifx \showISBNx    \undefined \def \showISBNx     #1{\unskip}     \fi
\ifx \showISBNxiii \undefined \def \showISBNxiii  #1{\unskip}     \fi
\ifx \showISSN     \undefined \def \showISSN      #1{\unskip}     \fi
\ifx \showLCCN     \undefined \def \showLCCN      #1{\unskip}     \fi
\ifx \shownote     \undefined \def \shownote      #1{#1}          \fi
\ifx \showarticletitle \undefined \def \showarticletitle #1{#1}   \fi
\ifx \showURL      \undefined \def \showURL       {\relax}        \fi
\providecommand\bibfield[2]{#2}
\providecommand\bibinfo[2]{#2}
\providecommand\natexlab[1]{#1}
\providecommand\showeprint[2][]{arXiv:#2}

\bibitem[Akbari et~al\mbox{.}(2021)]%
        {akbari2021vatt}
\bibfield{author}{\bibinfo{person}{Mohammadreza Akbari} {et~al\mbox{.}}} \bibinfo{year}{2021}\natexlab{}.
\newblock \showarticletitle{VATT: Transformers for Multimodal Self-Supervised Learning from Raw Video, Audio and Text}. In \bibinfo{booktitle}{\emph{Proceedings of the Advances in Neural Information Processing Systems (NeurIPS)}}.
\newblock


\bibitem[Burkhardt et~al\mbox{.}(2023)]%
        {burkhardt2023speech}
\bibfield{author}{\bibinfo{person}{Felix Burkhardt}, \bibinfo{person}{Johannes Wagner}, \bibinfo{person}{Hagen Wierstorf}, \bibinfo{person}{Florian Eyben}, {and} \bibinfo{person}{Bj{\"o}rn Schuller}.} \bibinfo{year}{2023}\natexlab{}.
\newblock \showarticletitle{Speech-based Age and Gender Prediction with Transformers}. In \bibinfo{booktitle}{\emph{Speech Communication; 15th ITG Conference}}. VDE, \bibinfo{pages}{46--50}.
\newblock


\bibitem[Cheng et~al\mbox{.}(2020)]%
        {10.1145/3394171.3413710}
\bibfield{author}{\bibinfo{person}{Kai Cheng}, \bibinfo{person}{Xin Liu}, \bibinfo{person}{Yiu-ming Cheung}, \bibinfo{person}{Rui Wang}, \bibinfo{person}{Xing Xu}, {and} \bibinfo{person}{Bineng Zhong}.} \bibinfo{year}{2020}\natexlab{}.
\newblock \showarticletitle{Hearing like Seeing: Improving Voice-Face Interactions and Associations via Adversarial Deep Semantic Matching Network}. In \bibinfo{booktitle}{\emph{Proceedings of the 28th ACM International Conference on Multimedia}} (Seattle, WA, USA) \emph{(\bibinfo{series}{MM '20})}. \bibinfo{publisher}{Association for Computing Machinery}, \bibinfo{address}{New York, NY, USA}, \bibinfo{pages}{448–455}.
\newblock
\showISBNx{9781450379885}
\urldef\tempurl%
\url{https://doi.org/10.1145/3394171.3413710}
\showDOI{\tempurl}


\bibitem[Hannun et~al\mbox{.}(2014)]%
        {hannun2014deep}
\bibfield{author}{\bibinfo{person}{Awni Hannun} {et~al\mbox{.}}} \bibinfo{year}{2014}\natexlab{}.
\newblock \showarticletitle{Deep Speech: Scaling up end-to-end speech recognition}.
\newblock \bibinfo{journal}{\emph{arXiv preprint arXiv:1412.5567}} (\bibinfo{year}{2014}).
\newblock


\bibitem[Horiguchi et~al\mbox{.}(2018)]%
        {10.1145/3240508.3240601}
\bibfield{author}{\bibinfo{person}{Shota Horiguchi}, \bibinfo{person}{Naoyuki Kanda}, {and} \bibinfo{person}{Kenji Nagamatsu}.} \bibinfo{year}{2018}\natexlab{}.
\newblock \showarticletitle{Face-Voice Matching using Cross-modal Embeddings}. In \bibinfo{booktitle}{\emph{Proceedings of the 26th ACM International Conference on Multimedia}} (Seoul, Republic of Korea) \emph{(\bibinfo{series}{MM '18})}. \bibinfo{publisher}{Association for Computing Machinery}, \bibinfo{address}{New York, NY, USA}, \bibinfo{pages}{1011–1019}.
\newblock
\showISBNx{9781450356657}
\urldef\tempurl%
\url{https://doi.org/10.1145/3240508.3240601}
\showDOI{\tempurl}


\bibitem[Kim et~al\mbox{.}(2019)]%
        {kim2019learning}
\bibfield{author}{\bibinfo{person}{Changil Kim}, \bibinfo{person}{Hijung~Valentina Shin}, \bibinfo{person}{Tae-Hyun Oh}, \bibinfo{person}{Alexandre Kaspar}, \bibinfo{person}{Mohamed Elgharib}, {and} \bibinfo{person}{Wojciech Matusik}.} \bibinfo{year}{2019}\natexlab{}.
\newblock \showarticletitle{On learning associations of faces and voices}. In \bibinfo{booktitle}{\emph{Computer Vision--ACCV 2018: 14th Asian Conference on Computer Vision, Perth, Australia, December 2--6, 2018, Revised Selected Papers, Part V 14}}. Springer, \bibinfo{pages}{276--292}.
\newblock


\bibitem[Nagrani et~al\mbox{.}(2018a)]%
        {10.1007/978-3-030-01261-8_5}
\bibfield{author}{\bibinfo{person}{Arsha Nagrani}, \bibinfo{person}{Samuel Albanie}, {and} \bibinfo{person}{Andrew Zisserman}.} \bibinfo{year}{2018}\natexlab{a}.
\newblock \showarticletitle{Learnable PINs: Cross-modal Embeddings for Person Identity}. In \bibinfo{booktitle}{\emph{Computer Vision – ECCV 2018: 15th European Conference, Munich, Germany, September 8-14, 2018, Proceedings, Part XIII}} (Munich, Germany). \bibinfo{publisher}{Springer-Verlag}, \bibinfo{address}{Berlin, Heidelberg}, \bibinfo{pages}{73–89}.
\newblock
\showISBNx{978-3-030-01260-1}
\urldef\tempurl%
\url{https://doi.org/10.1007/978-3-030-01261-8_5}
\showDOI{\tempurl}


\bibitem[Nagrani et~al\mbox{.}(2018b)]%
        {8578977}
\bibfield{author}{\bibinfo{person}{Arsha Nagrani}, \bibinfo{person}{Samuel Albanie}, {and} \bibinfo{person}{Andrew Zisserman}.} \bibinfo{year}{2018}\natexlab{b}.
\newblock \showarticletitle{Seeing Voices and Hearing Faces: Cross-Modal Biometric Matching}. In \bibinfo{booktitle}{\emph{2018 IEEE/CVF Conference on Computer Vision and Pattern Recognition}}. \bibinfo{pages}{8427--8436}.
\newblock
\urldef\tempurl%
\url{https://doi.org/10.1109/CVPR.2018.00879}
\showDOI{\tempurl}


\bibitem[Nawaz et~al\mbox{.}(2019)]%
        {8945863}
\bibfield{author}{\bibinfo{person}{Shah Nawaz}, \bibinfo{person}{Muhammad~Kamran Janjua}, \bibinfo{person}{Ignazio Gallo}, \bibinfo{person}{Arif Mahmood}, {and} \bibinfo{person}{Alessandro Calefati}.} \bibinfo{year}{2019}\natexlab{}.
\newblock \showarticletitle{Deep Latent Space Learning for Cross-Modal Mapping of Audio and Visual Signals}. In \bibinfo{booktitle}{\emph{2019 Digital Image Computing: Techniques and Applications (DICTA)}}. \bibinfo{pages}{1--7}.
\newblock
\urldef\tempurl%
\url{https://doi.org/10.1109/DICTA47822.2019.8945863}
\showDOI{\tempurl}


\bibitem[Nawaz et~al\mbox{.}(2021)]%
        {Nawaz2021CrossModal}
\bibfield{author}{\bibinfo{person}{Shah Nawaz}, \bibinfo{person}{Muhammad~Saad Saeed}, \bibinfo{person}{Pietro Morerio}, \bibinfo{person}{Arif Mahmood}, \bibinfo{person}{Ignazio Gallo}, \bibinfo{person}{Muhammad~Haroon Yousaf}, {and} \bibinfo{person}{Alessio Del~Bue}.} \bibinfo{year}{2021}\natexlab{}.
\newblock \showarticletitle{Cross-modal speaker verification and recognition: A multilingual perspective}. In \bibinfo{booktitle}{\emph{Proceedings of the IEEE/CVF Conference on Computer Vision and Pattern Recognition (CVPR)}}. \bibinfo{pages}{1682--1691}.
\newblock
\urldef\tempurl%
\url{https://doi.org/10.1109/CVPR.2021.00172}
\showDOI{\tempurl}


\bibitem[Ning et~al\mbox{.}(2022)]%
        {9400909}
\bibfield{author}{\bibinfo{person}{Hailong Ning}, \bibinfo{person}{Xiangtao Zheng}, \bibinfo{person}{Xiaoqiang Lu}, {and} \bibinfo{person}{Yuan Yuan}.} \bibinfo{year}{2022}\natexlab{}.
\newblock \showarticletitle{Disentangled Representation Learning for Cross-Modal Biometric Matching}.
\newblock \bibinfo{journal}{\emph{IEEE Transactions on Multimedia}}  \bibinfo{volume}{24} (\bibinfo{year}{2022}), \bibinfo{pages}{1763--1774}.
\newblock
\showISSN{1941-0077}
\urldef\tempurl%
\url{https://doi.org/10.1109/TMM.2021.3071243}
\showDOI{\tempurl}


\bibitem[Parkhi et~al\mbox{.}(2015)]%
        {parkhi2015}
\bibfield{author}{\bibinfo{person}{Omkar~M Parkhi}, \bibinfo{person}{Andrea Vedaldi}, {and} \bibinfo{person}{Andrew Zisserman}.} \bibinfo{year}{2015}\natexlab{}.
\newblock \showarticletitle{Deep face recognition}.
\newblock  (\bibinfo{year}{2015}).
\newblock


\bibitem[Ren et~al\mbox{.}(2024)]%
        {ren2024monocular}
\bibfield{author}{\bibinfo{person}{Xingyu Ren}, \bibinfo{person}{Jiankang Deng}, \bibinfo{person}{Yuhao Cheng}, \bibinfo{person}{Jia Guo}, \bibinfo{person}{Chao Ma}, \bibinfo{person}{Yichao Yan}, \bibinfo{person}{Wenhan Zhu}, {and} \bibinfo{person}{Xiaokang Yang}.} \bibinfo{year}{2024}\natexlab{}.
\newblock \showarticletitle{Monocular Identity-Conditioned Facial Reflectance Reconstruction}. In \bibinfo{booktitle}{\emph{Proceedings of the IEEE/CVF Conference on Computer Vision and Pattern Recognition (CVPR)}}.
\newblock


\bibitem[Saeed et~al\mbox{.}(2022)]%
        {saeed2022fusion}
\bibfield{author}{\bibinfo{person}{Muhammad~Saad Saeed}, \bibinfo{person}{Muhammad~Haris Khan}, \bibinfo{person}{Shah Nawaz}, \bibinfo{person}{Muhammad~Haroon Yousaf}, {and} \bibinfo{person}{Alessio Del~Bue}.} \bibinfo{year}{2022}\natexlab{}.
\newblock \showarticletitle{Fusion and orthogonal projection for improved face-voice association}. In \bibinfo{booktitle}{\emph{ICASSP 2022-2022 IEEE International Conference on Acoustics, Speech and Signal Processing (ICASSP)}}. IEEE, \bibinfo{pages}{7057--7061}.
\newblock


\bibitem[Saeed et~al\mbox{.}(2023)]%
        {10097207}
\bibfield{author}{\bibinfo{person}{Muhammad~Saad Saeed}, \bibinfo{person}{Shah Nawaz}, \bibinfo{person}{Muhammad~Haris Khan}, \bibinfo{person}{Muhammad Zaigham~Zaheer}, \bibinfo{person}{Karthik Nandakumar}, \bibinfo{person}{Muhammad~Haroon Yousaf}, {and} \bibinfo{person}{Arif Mahmood}.} \bibinfo{year}{2023}\natexlab{}.
\newblock \showarticletitle{Single-branch Network for Multimodal Training}. In \bibinfo{booktitle}{\emph{ICASSP 2023 - 2023 IEEE International Conference on Acoustics, Speech and Signal Processing (ICASSP)}}. \bibinfo{pages}{1--5}.
\newblock
\urldef\tempurl%
\url{https://doi.org/10.1109/ICASSP49357.2023.10097207}
\showDOI{\tempurl}


\bibitem[Saeed et~al\mbox{.}(2024)]%
        {saeed2024fame}
\bibfield{author}{\bibinfo{person}{Muhammad~Saad Saeed}, \bibinfo{person}{Shah Nawaz}, \bibinfo{person}{Muhammad~Salman Tahir}, \bibinfo{person}{Rohan~Kumar Das}, \bibinfo{person}{Muhammad~Zaigham Zaheer}, \bibinfo{person}{Marta Moscati}, \bibinfo{person}{Markus Schedl}, \bibinfo{person}{Muhammad~Haris Khan}, \bibinfo{person}{Karthik Nandakumar}, {and} \bibinfo{person}{Muhammad~Haroon Yousaf}.} \bibinfo{year}{2024}\natexlab{}.
\newblock \showarticletitle{Face-voice Association in Multilingual Environments (FAME) Challenge 2024 Evaluation Plan}.
\newblock \bibinfo{journal}{\emph{CoRR}}  \bibinfo{volume}{abs/2404.09342} (\bibinfo{year}{2024}).
\newblock
\showeprint[arxiv]{2404.09342}
\urldef\tempurl%
\url{https://arxiv.org/abs/2404.09342}
\showURL{%
\tempurl}


\bibitem[Serengil and Ozpinar(2021)]%
        {serengil2021lightface}
\bibfield{author}{\bibinfo{person}{Sefik~Ilkin Serengil} {and} \bibinfo{person}{Alper Ozpinar}.} \bibinfo{year}{2021}\natexlab{}.
\newblock \showarticletitle{HyperExtended LightFace: A Facial Attribute Analysis Framework}. In \bibinfo{booktitle}{\emph{2021 International Conference on Engineering and Emerging Technologies (ICEET)}}. IEEE, \bibinfo{pages}{1--4}.
\newblock
\urldef\tempurl%
\url{https://doi.org/10.1109/ICEET53442.2021.9659697}
\showDOI{\tempurl}


\bibitem[Shah et~al\mbox{.}(2023)]%
        {10136626}
\bibfield{author}{\bibinfo{person}{Saqlain~Hussain Shah}, \bibinfo{person}{Muhammad~Saad Saeed}, \bibinfo{person}{Shah Nawaz}, {and} \bibinfo{person}{Muhammad~Haroon Yousaf}.} \bibinfo{year}{2023}\natexlab{}.
\newblock \showarticletitle{Speaker recognition in realistic scenario using multimodal data}. In \bibinfo{booktitle}{\emph{2023 3rd International Conference on Artificial Intelligence (ICAI)}}. IEEE, \bibinfo{pages}{209--213}.
\newblock


\bibitem[Tsai et~al\mbox{.}(2019)]%
        {tsai2019multimodal}
\bibfield{author}{\bibinfo{person}{Yao-Hung~Hubert Tsai} {et~al\mbox{.}}} \bibinfo{year}{2019}\natexlab{}.
\newblock \showarticletitle{Multimodal Transformer for Unaligned Multimodal Language Sequences}. In \bibinfo{booktitle}{\emph{Proceedings of the 57th Annual Meeting of the Association for Computational Linguistics (ACL)}}. \bibinfo{pages}{6558--6569}.
\newblock


\bibitem[Wang et~al\mbox{.}(2019b)]%
        {8794873}
\bibfield{author}{\bibinfo{person}{Rui Wang}, \bibinfo{person}{Huaibo Huang}, \bibinfo{person}{Xufeng Zhang}, \bibinfo{person}{Jixin Ma}, {and} \bibinfo{person}{Aihua Zheng}.} \bibinfo{year}{2019}\natexlab{b}.
\newblock \showarticletitle{A Novel Distance Learning for Elastic Cross-Modal Audio-Visual Matching}. In \bibinfo{booktitle}{\emph{2019 IEEE International Conference on Multimedia \& Expo Workshops (ICMEW)}}. \bibinfo{pages}{300--305}.
\newblock
\urldef\tempurl%
\url{https://doi.org/10.1109/ICMEW.2019.00-70}
\showDOI{\tempurl}


\bibitem[Wang et~al\mbox{.}(2019a)]%
        {Wang_2019_CVPR}
\bibfield{author}{\bibinfo{person}{Xun Wang}, \bibinfo{person}{Xintong Han}, \bibinfo{person}{Weilin Huang}, \bibinfo{person}{Dengke Dong}, {and} \bibinfo{person}{Matthew~R. Scott}.} \bibinfo{year}{2019}\natexlab{a}.
\newblock \showarticletitle{Multi-Similarity Loss With General Pair Weighting for Deep Metric Learning}. In \bibinfo{booktitle}{\emph{Proceedings of the IEEE/CVF Conference on Computer Vision and Pattern Recognition (CVPR)}}. \bibinfo{pages}{5022--5030}.
\newblock
\urldef\tempurl%
\url{https://openaccess.thecvf.com/content_CVPR_2019/html/Wang_Multi-Similarity_Loss_With_General_Pair_Weighting_for_Deep_Metric_Learning_CVPR_2019_paper.html}
\showURL{%
\tempurl}


\bibitem[Wen et~al\mbox{.}(2021)]%
        {wen2021seeking}
\bibfield{author}{\bibinfo{person}{Peisong Wen}, \bibinfo{person}{Qianqian Xu}, \bibinfo{person}{Yangbangyan Jiang}, \bibinfo{person}{Zhiyong Yang}, \bibinfo{person}{Yuan He}, {and} \bibinfo{person}{Qingming Huang}.} \bibinfo{year}{2021}\natexlab{}.
\newblock \showarticletitle{Seeking the shape of sound: An adaptive framework for learning voice-face association}. In \bibinfo{booktitle}{\emph{Proceedings of the IEEE/CVF conference on computer vision and pattern recognition}}. \bibinfo{pages}{16347--16356}.
\newblock


\bibitem[Wen et~al\mbox{.}(2018)]%
        {wen2018disjoint}
\bibfield{author}{\bibinfo{person}{Yandong Wen}, \bibinfo{person}{Mahmoud~Al Ismail}, \bibinfo{person}{Weiyang Liu}, \bibinfo{person}{Bhiksha Raj}, {and} \bibinfo{person}{Rita Singh}.} \bibinfo{year}{2018}\natexlab{}.
\newblock \showarticletitle{Disjoint mapping network for cross-modal matching of voices and faces}.
\newblock \bibinfo{journal}{\emph{arXiv preprint arXiv:1807.04836}} (\bibinfo{year}{2018}).
\newblock


\bibitem[Xie et~al\mbox{.}(2019)]%
        {xie2019}
\bibfield{author}{\bibinfo{person}{Weidi Xie}, \bibinfo{person}{Arsha Nagrani}, \bibinfo{person}{Joon~Son Chung}, {and} \bibinfo{person}{Andrew Zisserman}.} \bibinfo{year}{2019}\natexlab{}.
\newblock \showarticletitle{Utterance-level aggregation for speaker recognition in the wild}. In \bibinfo{booktitle}{\emph{ICASSP 2019-2019 IEEE International Conference on Acoustics, Speech and Signal Processing (ICASSP)}}. IEEE, \bibinfo{pages}{5791--5795}.
\newblock


\bibitem[Zheng et~al\mbox{.}(2022)]%
        {9320535}
\bibfield{author}{\bibinfo{person}{Aihua Zheng}, \bibinfo{person}{Menglan Hu}, \bibinfo{person}{Bo Jiang}, \bibinfo{person}{Yan Huang}, \bibinfo{person}{Yan Yan}, {and} \bibinfo{person}{Bin Luo}.} \bibinfo{year}{2022}\natexlab{}.
\newblock \showarticletitle{Adversarial-Metric Learning for Audio-Visual Cross-Modal Matching}.
\newblock \bibinfo{journal}{\emph{IEEE Transactions on Multimedia}}  \bibinfo{volume}{24} (\bibinfo{year}{2022}), \bibinfo{pages}{338--351}.
\newblock
\showISSN{1941-0077}
\urldef\tempurl%
\url{https://doi.org/10.1109/TMM.2021.3050089}
\showDOI{\tempurl}


\bibitem[Zhu et~al\mbox{.}(2022)]%
        {ijcai2022p526}
\bibfield{author}{\bibinfo{person}{Boqing Zhu}, \bibinfo{person}{Kele Xu}, \bibinfo{person}{Changjian Wang}, \bibinfo{person}{Zheng Qin}, \bibinfo{person}{Tao Sun}, \bibinfo{person}{Huaimin Wang}, {and} \bibinfo{person}{Yuxing Peng}.} \bibinfo{year}{2022}\natexlab{}.
\newblock \showarticletitle{Unsupervised Voice-Face Representation Learning by Cross-Modal Prototype Contrast}. In \bibinfo{booktitle}{\emph{Proceedings of the Thirty-First International Joint Conference on Artificial Intelligence, {IJCAI-22}}}, \bibfield{editor}{\bibinfo{person}{Lud~De Raedt}} (Ed.). \bibinfo{publisher}{International Joint Conferences on Artificial Intelligence Organization}, \bibinfo{pages}{3787--3794}.
\newblock
\urldef\tempurl%
\url{https://doi.org/10.24963/ijcai.2022/526}
\showDOI{\tempurl}
\newblock
\shownote{Main Track}.


\end{thebibliography}










\end{document}